\documentclass{article}

\usepackage{arxiv}

\usepackage[utf8]{inputenc} 
\usepackage[T1]{fontenc}    
\usepackage{hyperref}       
\usepackage{url}            
\usepackage{amsfonts}       
\usepackage{xcolor}         

\usepackage{subfigure}
\usepackage{pgf}
\usepackage{geometry}

\DeclareUnicodeCharacter{2212}{-}

\author{Valentin Guillet \\
       ISAE-SUPAERO, Université de Toulouse, France \\
       \texttt{valentin.guillet@isae-supaero.fr} \\
       \And
       Dennis Wilson \\
       ISAE-SUPAERO, Université de Toulouse, France \\
       \texttt{dennis.wilson@isae-supaero.fr} \\
       \And
       Carlos Aguilar-Melchor \\
       Sandbox AQ, France \\
       \texttt{carlos@sandboxaq.com} \\
       \And
       Emmanuel Rachelson \\
       ISAE-SUPAERO, Université de Toulouse, France \\
       \texttt{emmanuel.rachelson@isae-supaero.fr} \\
}

\title{On Neural Consolidation for Transfer in Reinforcement Learning}

\begin{document}

\maketitle

\let\thefootnote\relax\footnotetext{Work published at the IEEE Symposium on Adaptive Dynamic Programming and Reinforcement Learning (IEEE ADPRL), 2022}

\begin{abstract}
    Although transfer learning is considered to be a milestone in deep reinforcement learning, the mechanisms behind it are still poorly understood.
    In particular, predicting if knowledge can be transferred between two given tasks is still an unresolved problem.
    In this work, we explore the use of network distillation as a feature extraction method to better understand the context in which transfer can occur.
    Notably, we show that distillation does not prevent knowledge transfer, including when transferring from multiple tasks to a new one, and we compare these results with transfer without prior distillation.
    We focus our work on the Atari benchmark due to the variability between different games, but also to their similarities in terms of visual features.
\end{abstract}

\section{Introduction}
\label{sec:introduction}

In spite of the rapid progress made in Deep Reinforcement Learning (RL) \cite{sutton2018reinforcement} in the last decade, and although state-of-the-art algorithms are more and more efficient, many fundamental issues still have not been solved and remain major limitations in current approaches. In particular, existing algorithms train networks from scratch on each new task, which is very computationally costly.
This issue motivated the development of transfer learning \cite{pratt1991direct,taylor2009transfer}, the study of how to transfer and reuse knowledge from a neural network to another in order to accelerate learning and benefit from previously acquired abilities.
Various methods for transfer have been proposed recently, from plain fine-tuning, to a more complex use of distillation in a multi-task setting \cite{rusu2016policy}.

Although primarily developed for network compression \cite{bucilua2006model}, \emph{distillation} is a technique that aims at copying the behavior of a \emph{teacher} neural network into a \emph{student} one by ensuring the two represent the same function. It has been successfully used to compress multiple teachers in a single student, thus achieving multi-task learning \cite{hinton2015distilling}.
We investigate how distillation in a multi-task context, which we refer to as \emph{network consolidation}, is useful for knowledge transfer and can help understand the underlying mechanisms behind transfer.
Specifically, we compare different methods to achieve consolidation on multiple visual RL tasks and discuss the importance of key details in the algorithmic design.
We also show that transfer can occur even when the consolidation process does not reach convergence.
Finally, we argue that consolidation filters out networks that would lead to negative transfer, while preserving the benefits of positive transfer cases.

In order to study these claims, we propose an experimental protocol based on the use of the Actor-Mimic algorithm \cite{parisotto2016actor-mimic} that alternates between training and consolidation phases.
Our approach is motivated by current neurobiological theories modeling knowledge transfer and lifelong learning in the mammalian brain, such as the Complementary Learning Systems (CLS) theory \cite{mcclelland1995why}.
CLS states that memorization is based on two distinct parts of the brain: the hippocampus, responsible for short-term adaptation and rapid learning, and the neocortex, which assimilates this knowledge slowly and retains it on a long-term basis.

We present an overview of related works in Section \ref{sec:related_work}, before describing our experimental protocol in Section \ref{sec:amn_for_consolidation}.
Section \ref{sec:improving_performance_via_consolidation} discusses the effect of using the Actor-Mimic algorithm for consolidation, on the performance obtained within a given set of tasks.
Section \ref{sec:consolidation_for_transfer} focuses on knowledge transfer and generalization to new tasks.
Section \ref{sec:transfer_without_consolidation} provides a comparison baseline illustrating that consolidation mitigates the effects of negative transfer. We conclude in Section \ref{sec:conclu}.

\section{Background and Related Work}
\label{sec:related_work}

Distillation was originally proposed by \cite{rusu2016policy} and is among the most promising methods to achieve transfer between tasks.
\cite{parisotto2016actor-mimic} and \cite{jung2016less-forgetting} extend the core idea of learning several functions as one, and add an incentive to also copy the features in order to guide the training process.
Similarly, \cite{teh2017distral} builds a central network, encoding common behaviors, to share knowledge between tasks.

One major challenge in RL today is lifelong learning, i.e. how to solve different tasks sequentially while avoiding \emph{catastrophic forgetting}. Different approaches exist to tackle this problem. We follow the division in three categories proposed in \cite{parisi2019continual}.
One possibility is to periodically modify the network architecture when facing new tasks in order to enhance its representative power \cite{yoon2018lifelong,rusu2016progressive,fernando2017pathnet}.
Another approach is to use regularization to preserve previously acquired knowledge \cite{li2018learning,kirkpatrick2017overcoming,zenke2017continual}.
Finally, the lifelong learning problem can be reduced to a multi-task learning one by using a rehearsal strategy, memorizing every task encountered \cite{lopez-paz2017gradient,rebuffi2017icarl,kaiser2020model,ha2018world,shin2017continual}.
These three main categories are not mutually exclusive, and many of these algorithms make use of different techniques that belong to two categories.

The idea of alternating between an active phase of pattern-separated learning and a passive phase of generalization as inspired by CLS theory has also been explored before. In particular, \cite{berseth2018progressive} introduces the PLAID algorithm that progressively grows a central network using distillation on newly encountered tasks.
Similarly, \cite{schwarz2018progress} successively compresses different expert networks in a \emph{knowledge base} that is then reused by new experts via lateral layer-wise connections \cite{rusu2016progressive}. \cite{blakeman2020complementary} introduces a Self-Organising Map to DRL to simulate complementary learning in a neocortical and a hippocampal system, improving learning on grid world control and demonstrating the biological plausibility of artificial CLS.

Instead of learning to solve multiple tasks, another possibility is to learn how to be efficient at learning: this is the meta-learning approach.
One intuitive way to achieve this is by using a meta-algorithm to output a set of neural network weights which are then used as initialization for solving new tasks \cite{finn2017model-agnostic}.
On the other hand, \cite{beaulieu2020learning} proposes the use of a second network whose role is to deactivate part of a typical neural network. By analogy with the human brain, this network is called the neuromodulatory network as it is responsible for activating or deactivating part of the main network depending on the current task to solve.
Finally, \cite{he2020task} proposes a framework for meta-algorithms which divides them into a ``What'' part whose objective is to identify the current running task from context data, and a ``How'' part responsible for producing a set of parameters for a neural network that will be able to solve this task.

\section{Actor-Mimic Networks for consolidation in Lifelong Learning}
\label{sec:amn_for_consolidation}

In order to study the consolidation process and its interaction with knowledge transfer, we explore the use of the Actor-Mimic (Network) algorithm \cite[AMN]{parisotto2016actor-mimic} that acts as a policy distillation algorithm with an additional incentive to imitate the teacher's features.
In standard policy distillation, as proposed by \cite{rusu2016policy}, the distilled network --- also called student network --- learns to reproduce the output of multiple expert networks (policy regression objective) using supervised learning.
In addition, the AMN algorithm adds another feature regression objective that regularizes the features of the student network (defined as the outputs of the second-to-last layer) towards the features of the experts.
Intuitively, the policy regression objective teaches the student \emph{how} it should act while this feature regression objective teaches the result of the expert's ``thinking process'' that indicates \emph{why} it should act that way.

The AMN algorithm makes it possible to consolidate several expert networks at the same time while extracting features containing the same information as the experts.
As the input states of target tasks can be quite different in nature (e.g. graphical features, color palette, etc.), it is a desirable property for the extracted features in the consolidated network to represent abstract concepts that facilitate generalization across tasks.
To evaluate this property, we propose a new training protocol composed of two phases that emulate day-night cycles: an active learning phase in which neural networks ---coined ``expert networks''--- are trained individually on a set of visual RL tasks, and a passive imitation phase in which the knowledge acquired by all experts is consolidated into a central AMN that retains knowledge in the long term.

During the active phase, each expert network is trained on its corresponding task using a standard RL algorithm.
We use Rainbow \cite{hessel2018rainbow} in the present study, as implemented in the Dopamine framework \cite{castro2018dopamine}, with the typical architecture of 3 convolutional layers followed by two dense layers as introduced by \cite{mnih2015human}.
These phases are interrupted early, before performance reaches state-of-the-art levels, as the objective is to extract general features that will encourage the next experts to avoid focusing on task-specific pixel-level characteristics.
The passive phase consists in consolidating an AMN from these experts.
We use the same hyperparameters as in the original AMN work \cite{parisotto2016actor-mimic}.
We keep one prioritized replay buffer \cite{schaul2015prioritized} per task during the passive phase, which is filled by interacting with the tasks using the AMN's actions.
To enforce exploration, we use an $\epsilon$-greedy policy both for the experts and the AMNs, with $\epsilon$ starting from $1$ and annealing progressively to $0.1$.
Imitation is less sample-demanding than the RL policy optimization process \cite{rusu2016policy}, therefore the AMN only needs a fraction of the number of time steps of the active phase to exhibit the same performance as the experts.
The next active phase is then started by initializing new experts with the AMN weights, a process we discuss specifically in Section \ref{subsec:transfer_passive_active}.

We evaluate this generic protocol on the Atari benchmark \cite{bellemare2013arcade}, and more precisely on the games Breakout, Carnival, Pong, SpaceInvaders and VideoPinball, selected for their diversity and their balanced difficulty.
Although the choice of these specific games may limit our analysis, this benchmark is representative in that some of these games can appear to human players as similar (e.g. hit a ball moving in straight lines with a paddle) but are different from a visual perspective.
As recommended by \cite{castro2018dopamine} we report the training performance of the experts by averaging the return on every completed episode during iterations of 50000 time steps.
For the sake of readability, we only report here the most illustrative results.
Our code and complete results are available at \url{https://anonymous.4open.science/r/ConsolidationForTransferInRL}.

\section{Improving performance via consolidation}
\label{sec:improving_performance_via_consolidation}


\subsection{On the passive consolidation phase}
\label{subsec:passive_consolidation_phase}

\begin{figure*}
    \centering
    \subfigure {
        \resizebox{0.33\textwidth}{!}{\includegraphics{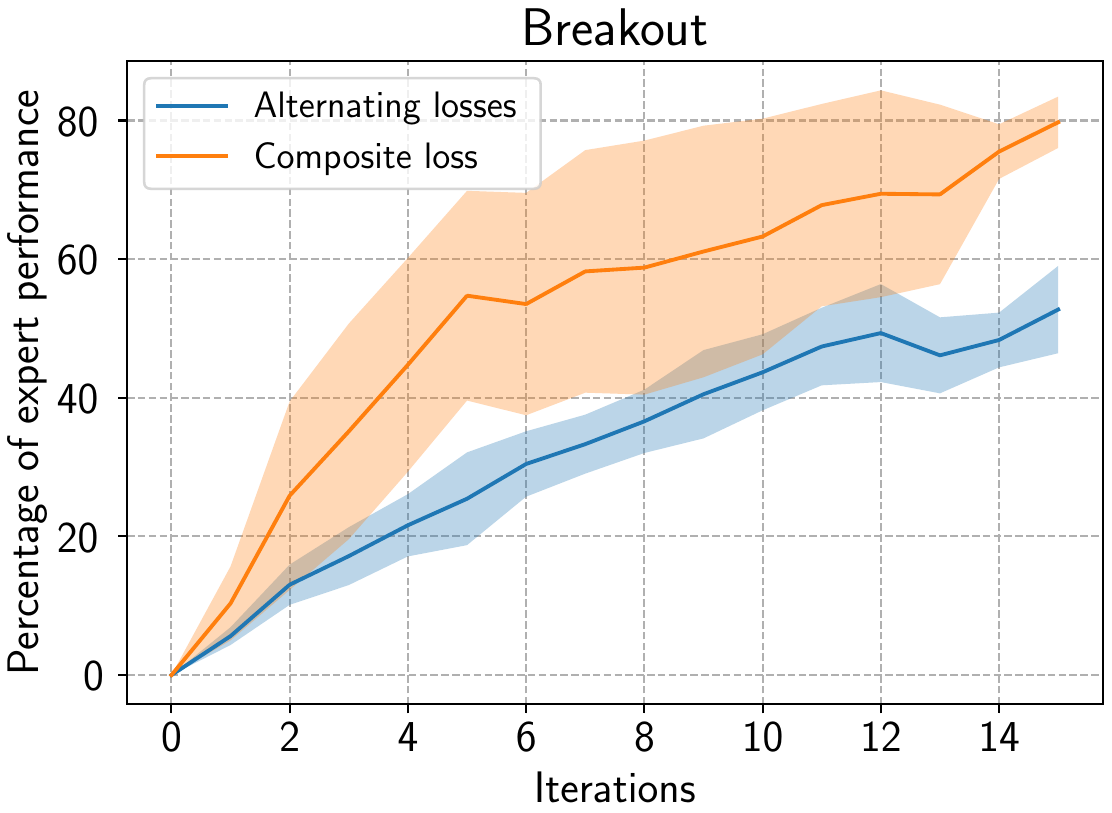}}
        \label{fig:sum_of_losses_breakout}
    }
    \subfigure {
        \resizebox{0.30\textwidth}{!}{\includegraphics{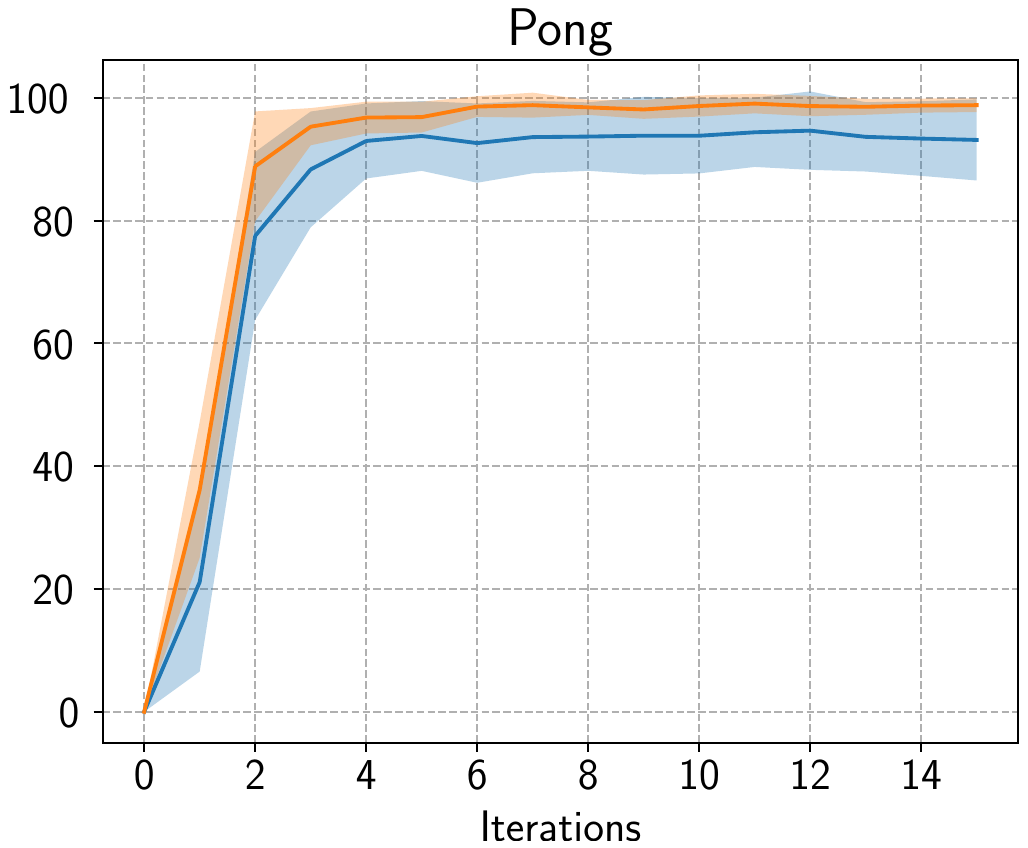}}
        \label{fig:sum_of_losses_pong}
    }
    \subfigure {
        \resizebox{0.30\textwidth}{!}{\includegraphics{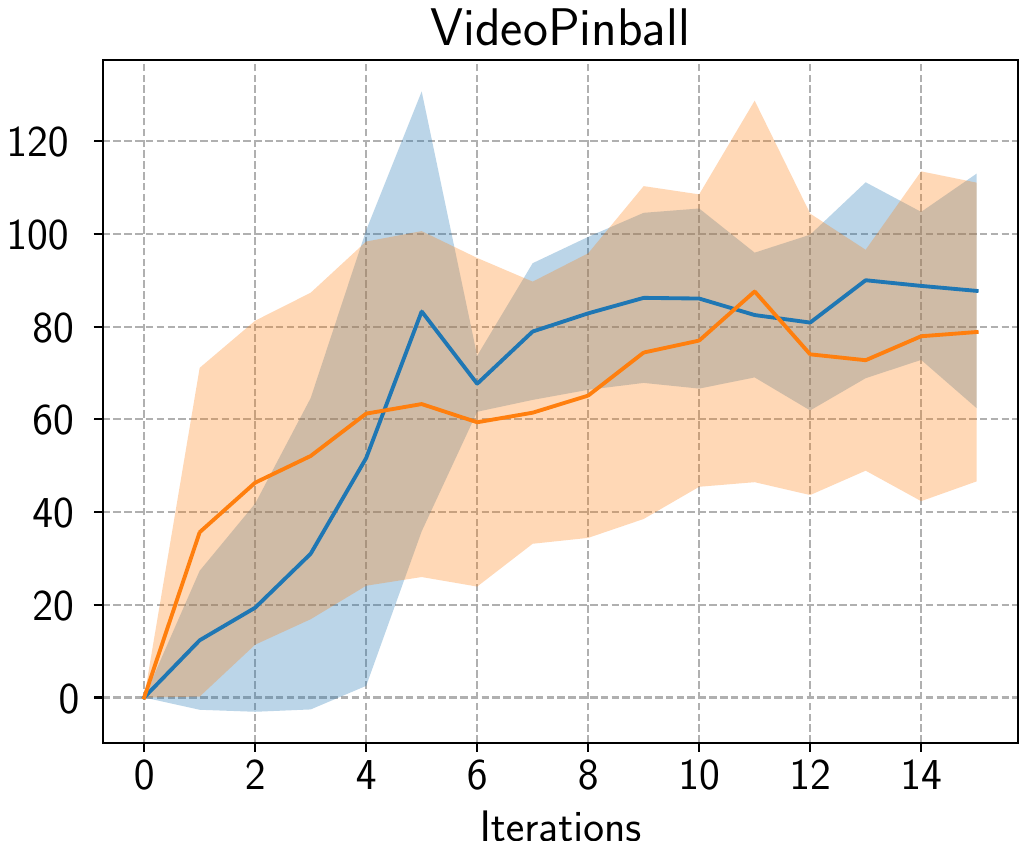}}
        \label{fig:sum_of_losses_videopinball}
    }

    \caption{Percentage of the experts final score for an AMN consolidated on 5 experts. Composite loss vs. alternating individual losses. Each experiment repeated 3 times (shaded area=1$\sigma$).}
    \label{fig:sum_of_losses}
\end{figure*}

The AMN is trained on samples issued by the experts on their respective tasks. The scheduling of these tasks is an important design choice.
A straightforward approach consists in minimizing a single composite loss that sums the losses for each expert network.
Then, one gradient descent step minimizes all individual losses at the same time, ensuring the simultaneity.
However, two tasks could theoretically result in opposite gradient directions that would cancel one another, preventing the consolidated network to improve on any of these tasks. This issue was raised by \cite{yu2020gradient} who shows that this situation can occur frequently.
In our study, this issue did not occur enough to degrade performance significantly and the use of a single composite loss gave satisfying results.
\figurename \ref{fig:sum_of_losses} compares the performance when optimizing a single composite loss and when optimizing the separate losses when switching tasks at the end of each episode.

\begin{figure*}
    \centering
    \subfigure {
        \resizebox{0.31\textwidth}{!}{\includegraphics{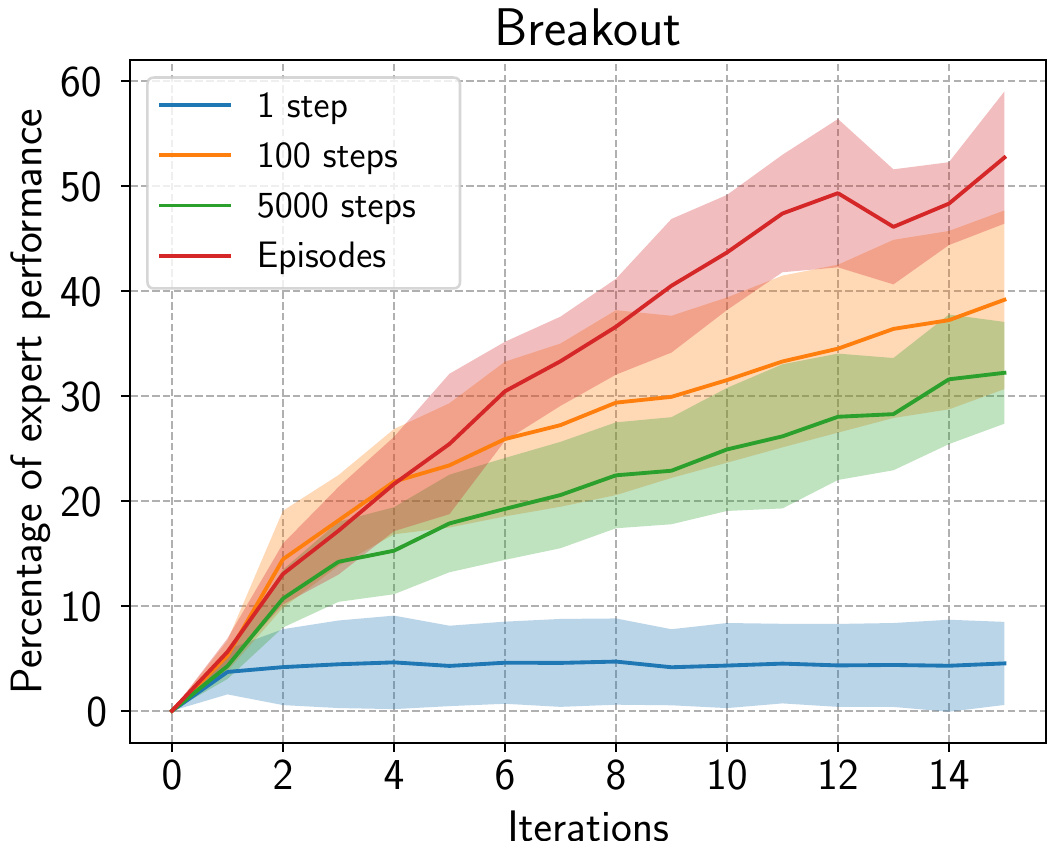}}
        \label{fig:by_steps_breakout}
    }
    \subfigure {
        \resizebox{0.31\textwidth}{!}{\includegraphics{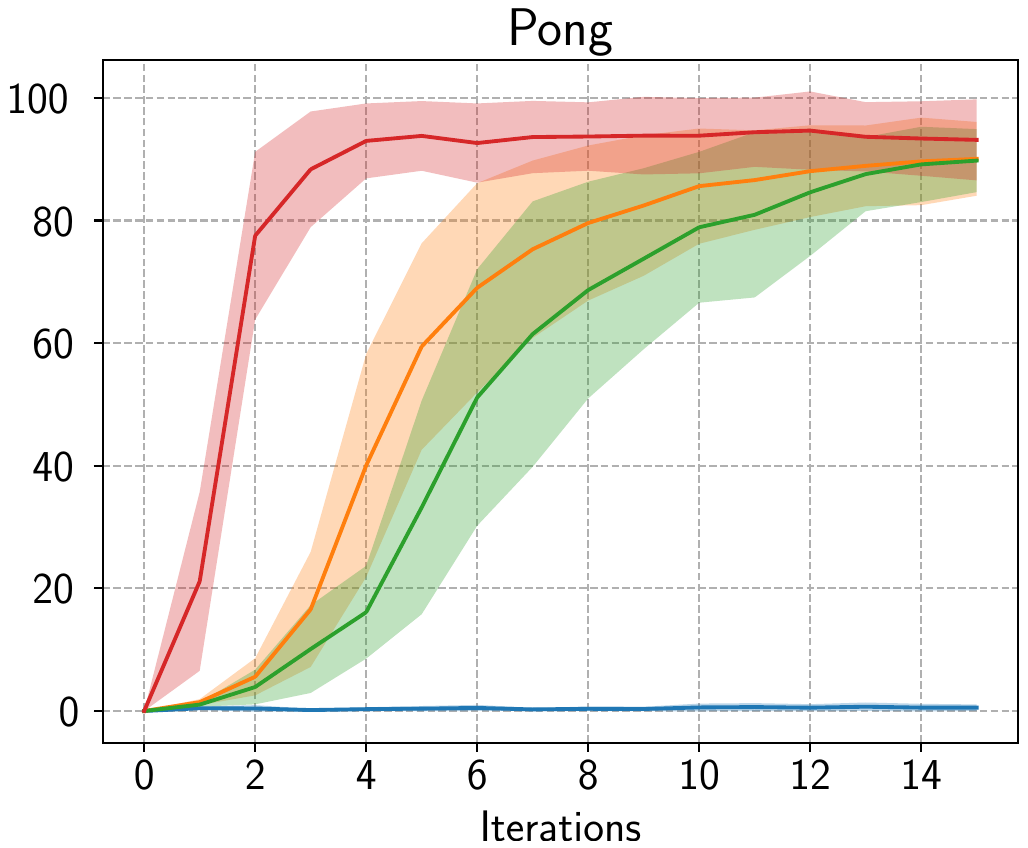}}
        \label{fig:by_steps_pong}
    }
    \subfigure {
        \resizebox{0.31\textwidth}{!}{\includegraphics{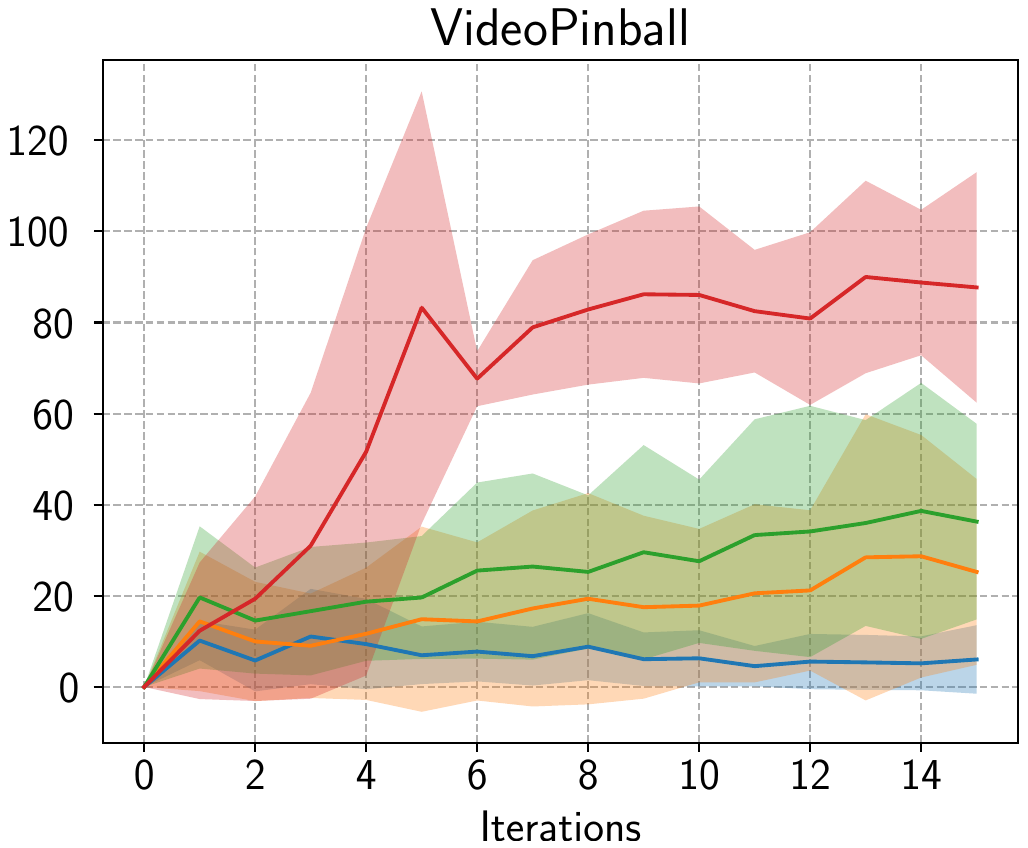}}
        \label{fig:by_steps_videopinball}
    }

    \caption{Percentage of the experts final score for an AMN. Switching between tasks every 1, 100, 5000 steps or every episode.}
    \label{fig:by_steps}
\end{figure*}

Instead, one can alternate between tasks during consolidation and optimize the different losses sequentially.
This approach also solves the issue of losses of different orders of magnitude when used in conjunction with the Adam optimizer.
However, it requires tuning the frequency with which to switch from one task to the next.
\figurename \ref{fig:by_steps} compares four different values for this frequency: a high frequency switch (every time step), a medium frequency (every 100 time steps), a low frequency (every 5000 steps) and switching after every episode.
Switching at every time step prevented any learning, and the obtained weights are not a good initialization point for the next active phase, suggesting that no interesting features were extracted during training.
For both medium and low frequency switching, the results were task-dependent.
Finally, despite the imbalance of number of steps per episode across tasks, it appears that switching only at the end of full episodes results in better performance overall.

\subsection{Transfer from passive to active phase}
\label{subsec:transfer_passive_active}

To study the contribution of the passive phase, we consolidate an AMN on 5 games, then initialize an expert with the AMN weights and let it train on one of these 5 games.
For tasks that feature less actions than the number of AMN outputs, we simply copy the fitting subpart of the network.

\begin{figure*}
    \centering
    \subfigure {
        \resizebox{0.31\textwidth}{!}{\includegraphics{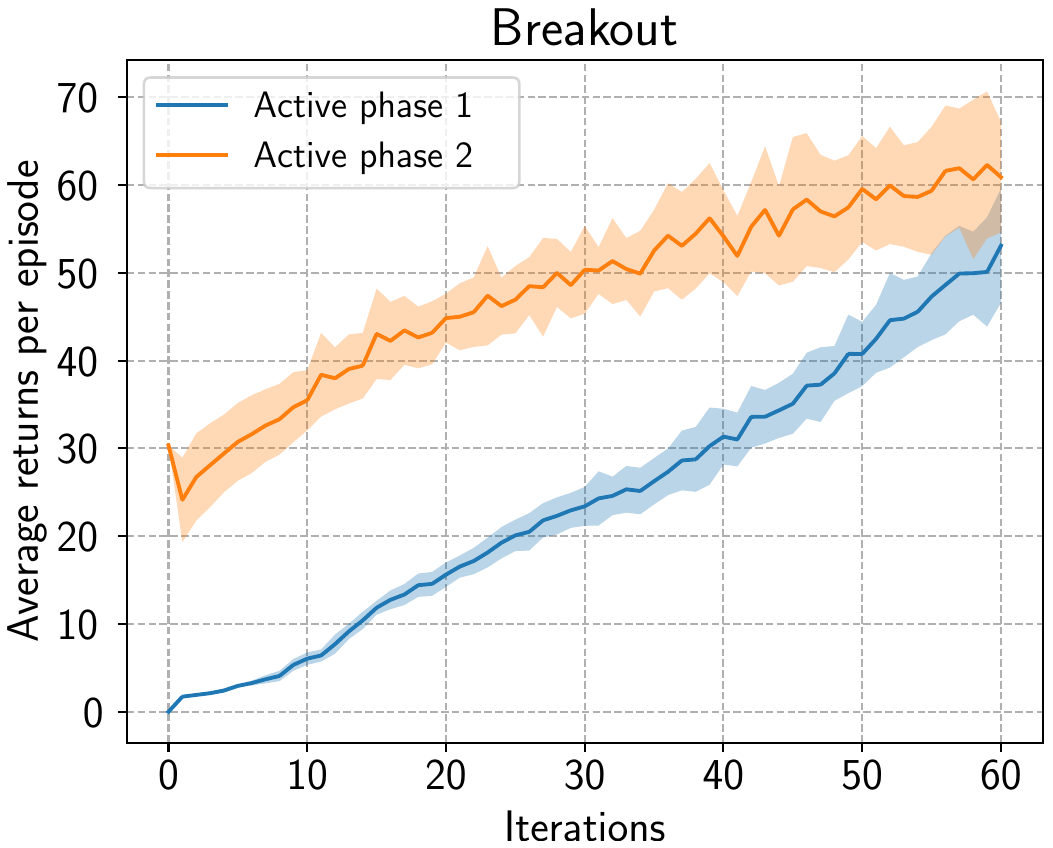}}
        \label{fig:passive_to_active_breakout}
    }
    \subfigure {
        \resizebox{0.31\textwidth}{!}{\includegraphics{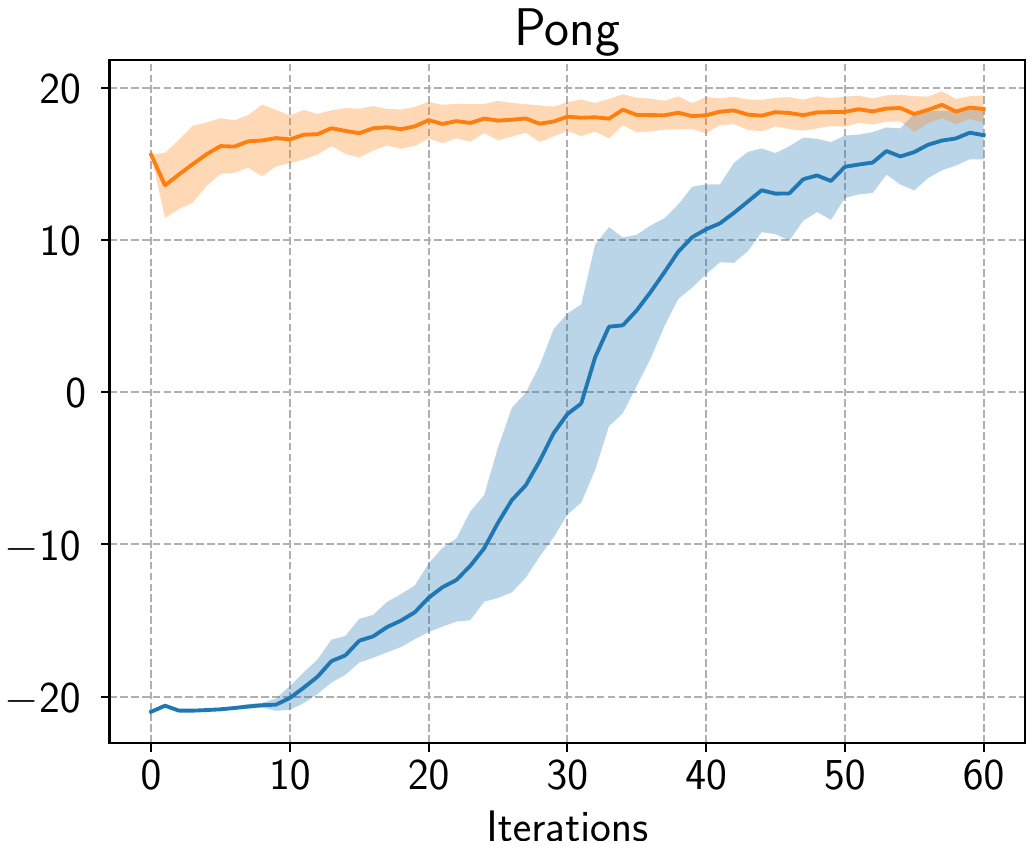}}
        \label{fig:passive_to_active_pong}
    }
    \subfigure {
        \resizebox{0.31\textwidth}{!}{\includegraphics{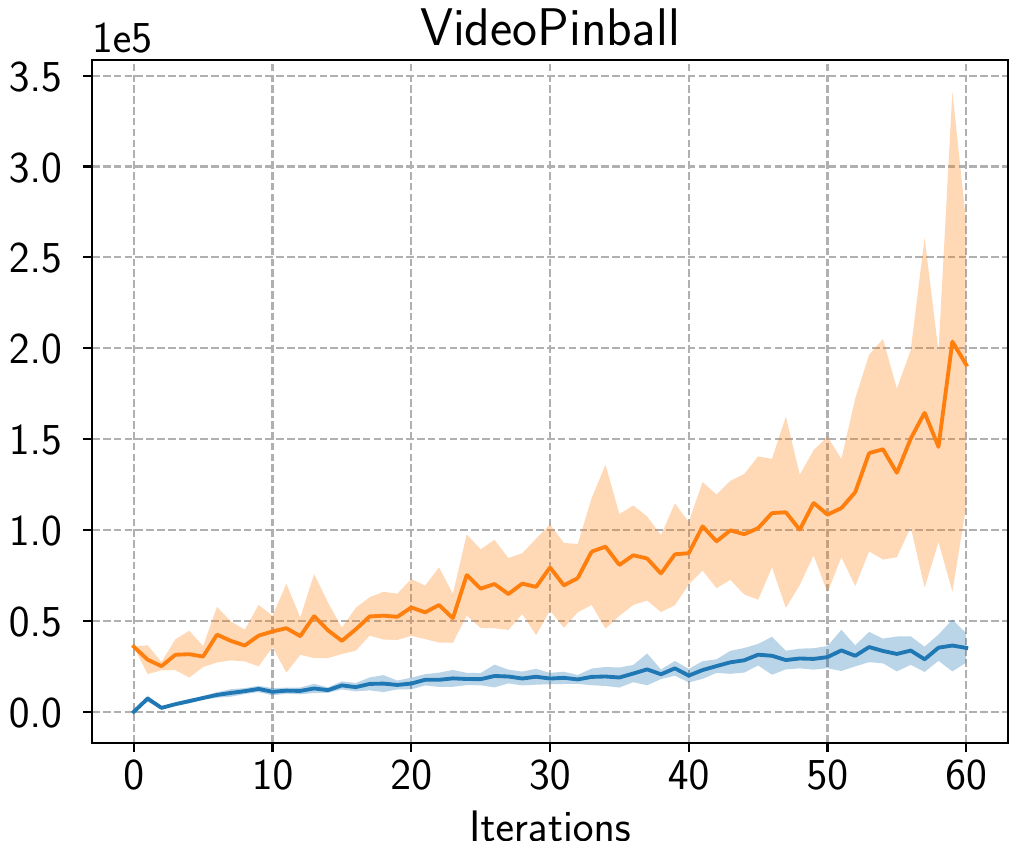}}
        \label{fig:passive_to_active_videopinball}
    }

    \caption{Randomly initialized expert vs. AMN initialization.}
    \label{fig:passive_to_active}
\end{figure*}

\figurename \ref{fig:passive_to_active} compares the average returns per episode between the expert network trained on the first active phase (thus initialized randomly) and the one trained on the second active phase (initialized with the AMN weights).
This experiment confirms that the transfer has a significant jumpstart effect as the second expert networks all start above the initial value of a random policy.
However, the long term effect of this initial performance boost is largely dependent on the task tackled, and sometimes it disappears quickly (Breakout) or keeps a certain advantage during the whole training (VideoPinball).

\begin{figure*}
    \centering
    \subfigure {
        \resizebox{0.31\textwidth}{!}{\includegraphics{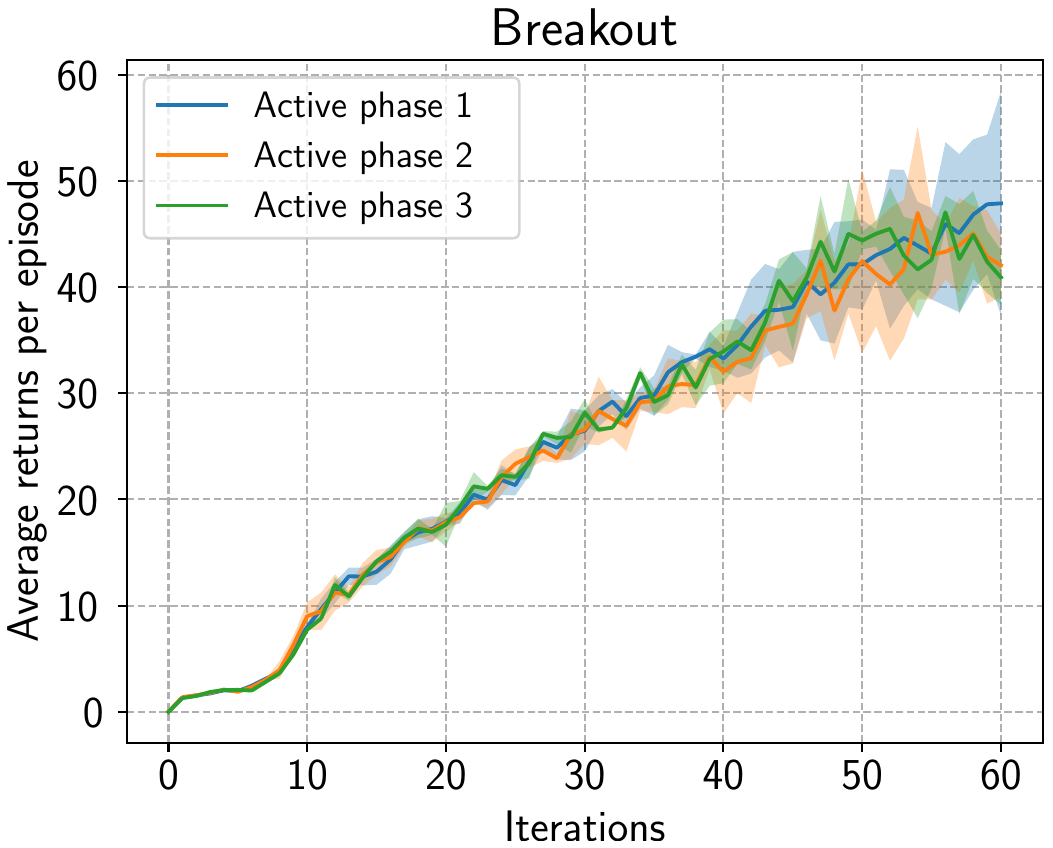}}
        \label{fig:feature_concatenation_results_breakout}
    }
    \subfigure {
        \resizebox{0.31\textwidth}{!}{\includegraphics{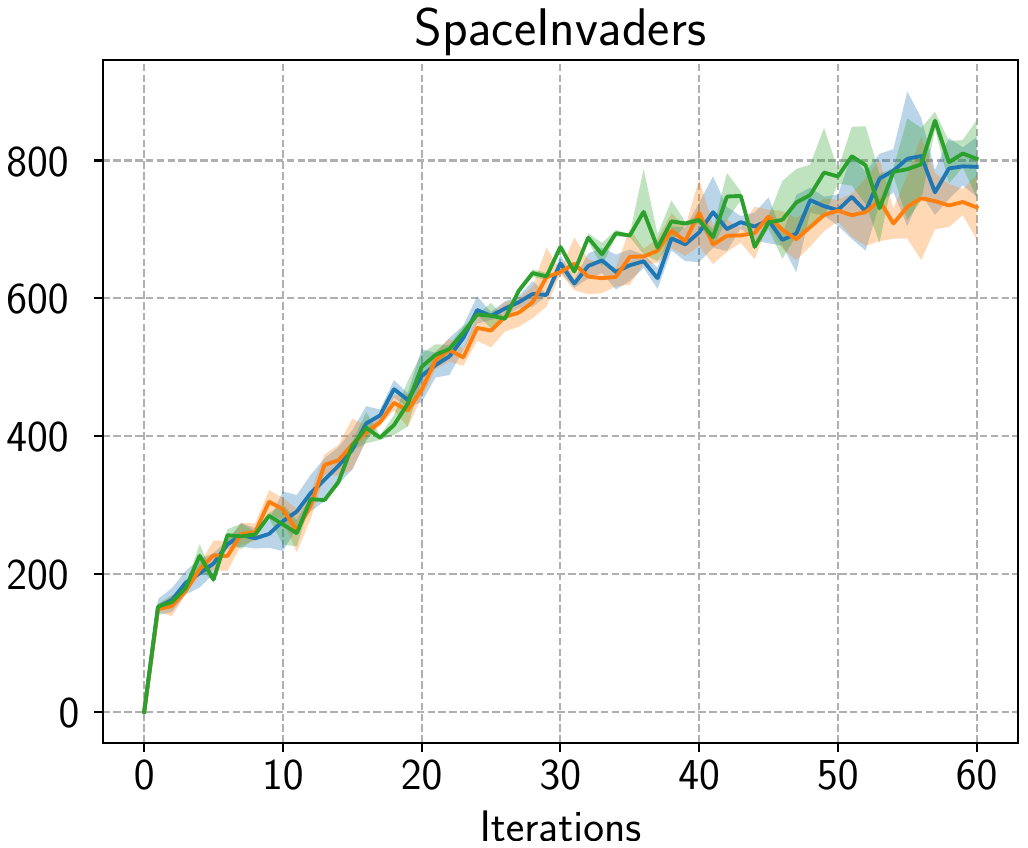}}
        \label{fig:feature_concatenation_results_spaceinvaders}
    }
    \subfigure {
        \resizebox{0.31\textwidth}{!}{\includegraphics{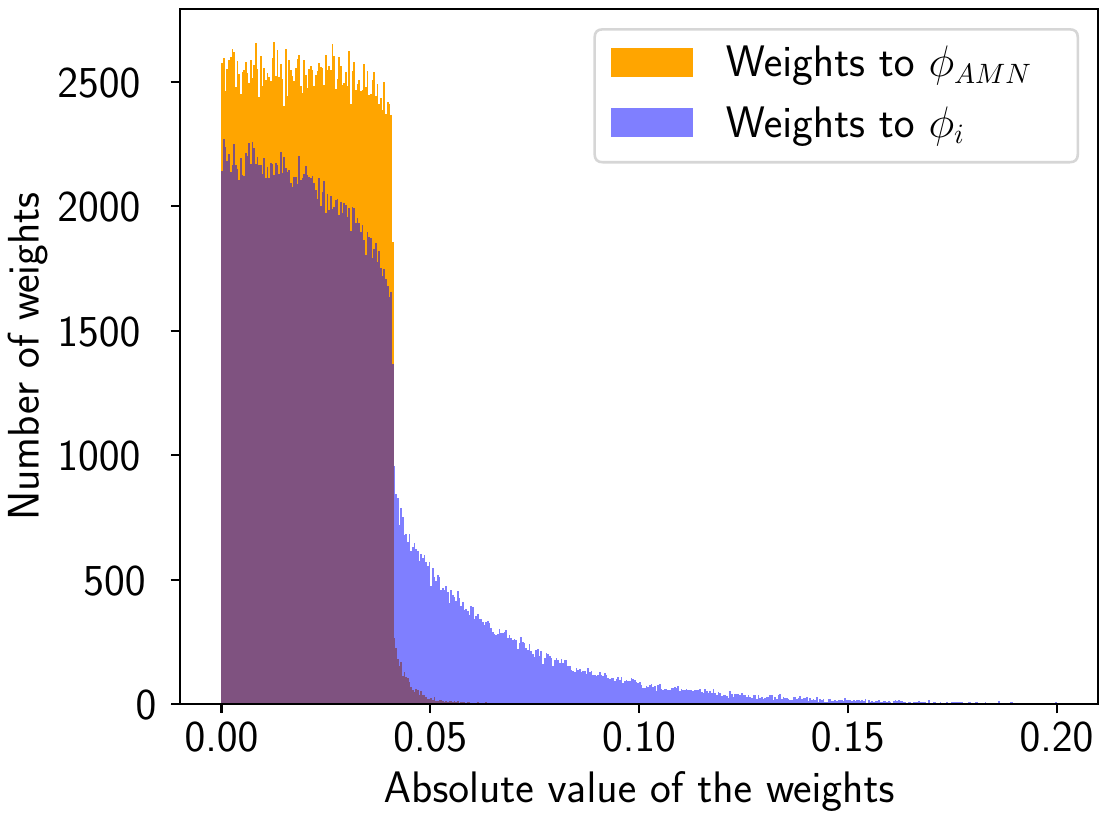}}
        \label{fig:feature_concatenation_weight_analysis}
    }

    \caption{Successive active phases for experts with lateral connections to the AMN.}
    \label{fig:feature_concatenation_results}
\end{figure*}

Not all extracted features are useful for all tasks and zero-weight connections are known to be beneficial to initializing neural networks.
Instead of forcing the feature initialization, a more elaborate approach is to use a default initialization for a new expert and make the AMN features accessible via a lateral connection from the AMN feature layer to the expert's output layer.
This flexible approach trades off the initial jumpstart effect to offer more degrees of freedom to the gradient descent process.
\figurename \ref{fig:feature_concatenation_results} shows the evolution of the average return per episode, for such experts, trained during successive active phases.
The similarity between training profiles and the comparison with \figurename \ref{fig:passive_to_active} suggests the features built during consolidation are not exploited in the expert's gradient descent process.
To confirm, we plot the histogram of the last layer weights absolute values and separate the weights stemming from the AMN from those initialized randomly.
The weights linking to the AMN features have very low magnitude, confirming that the gradient descent does not exploit the AMN features.
So plain copy of features is not sufficient for transfer in Deep RL, which seems rather to benefit from good initialization (confirming the analysis of \cite{finn2017model-agnostic}).

\section{Consolidation for Transfer Learning and Domain Generalization}
\label{sec:consolidation_for_transfer}

In the previous section, we analyzed the effects of consolidation for agents playing the \emph{same} set of games repeatedly, illustrating the conditions under which we observe jumpstart and asymptotic benefits.
In this section, we now train each active phase on a completely new set of tasks to assess whether consolidation permits generalization across tasks.

\subsection{Consolidation towards new unseen tasks}

\begin{figure*}
    \centering
    \subfigure {
        \resizebox{0.31\textwidth}{!}{\includegraphics{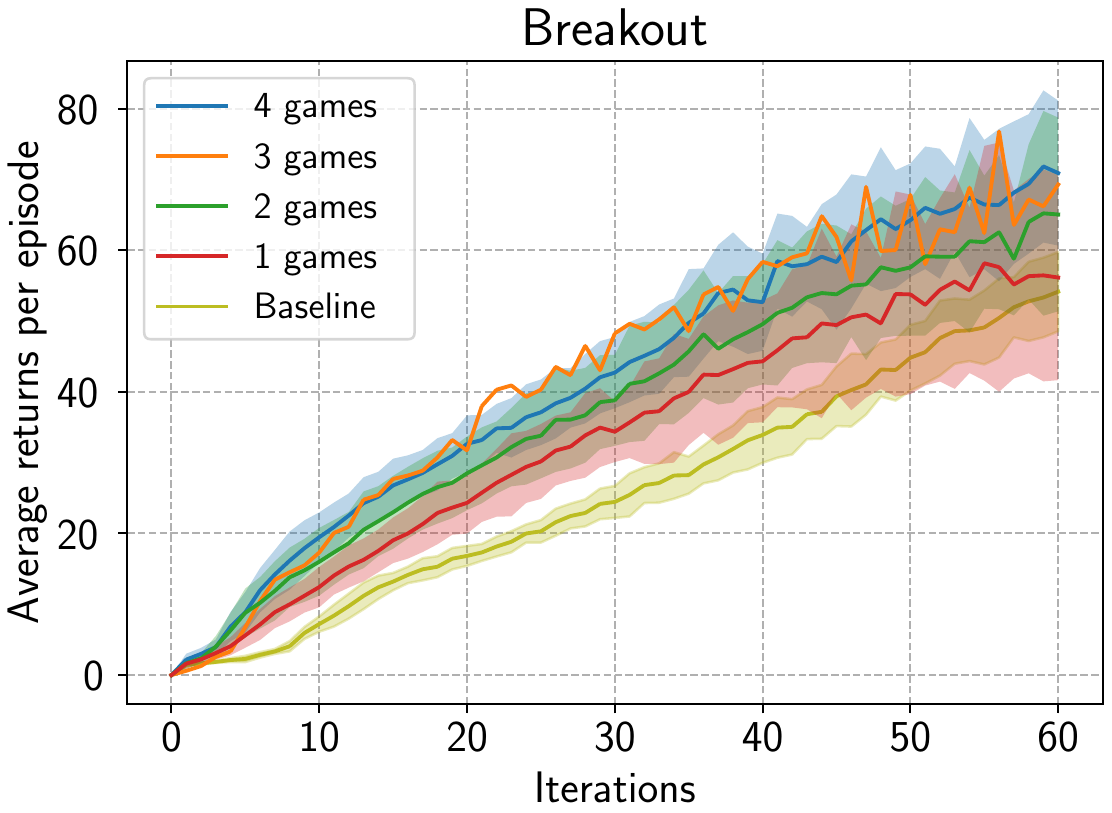}}
        \label{fig:comparison_nb_games_breakout}
    }
    \subfigure {
        \resizebox{0.31\textwidth}{!}{\includegraphics{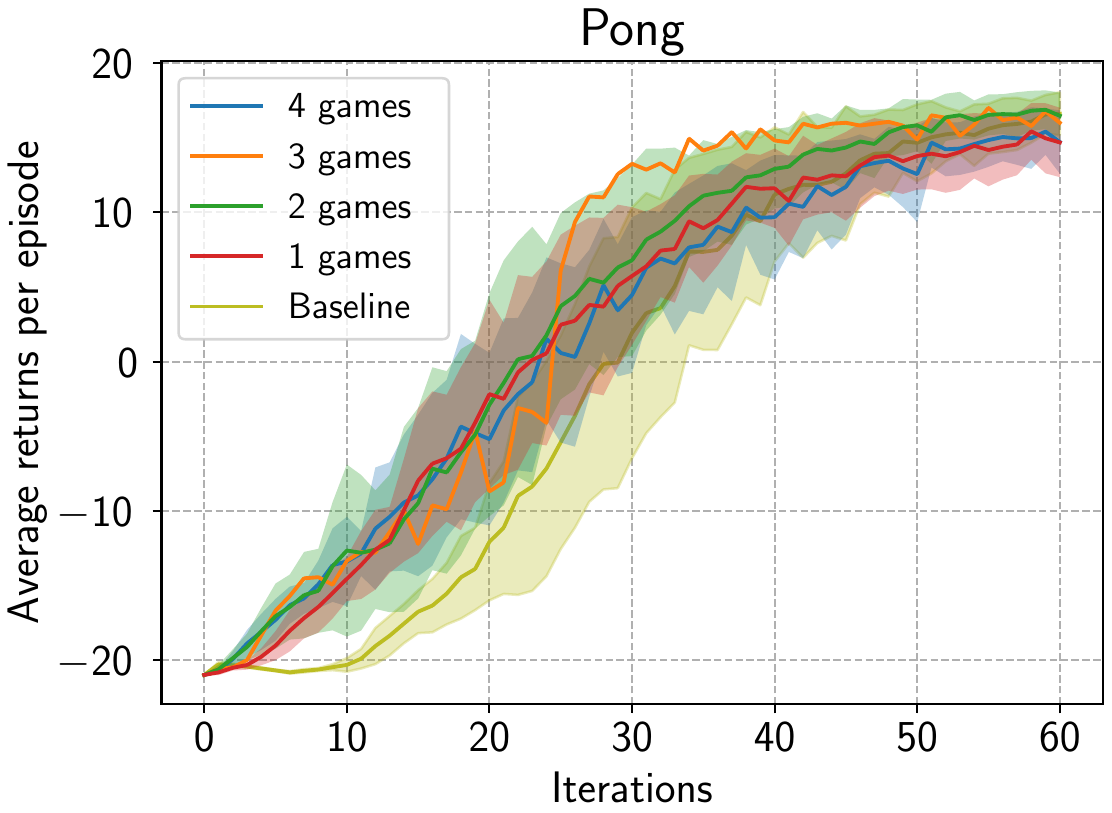}}
        \label{fig:comparison_nb_games_pong}
    }
    \subfigure {
        \resizebox{0.31\textwidth}{!}{\includegraphics{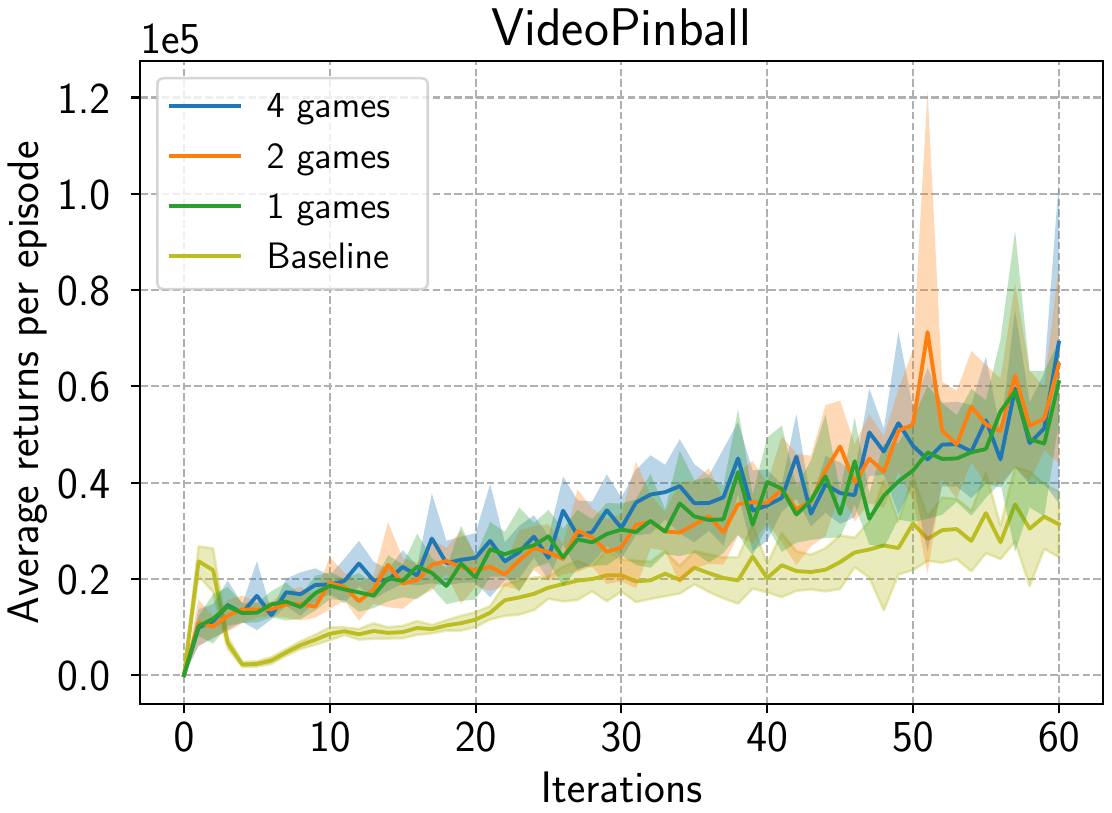}}
        \label{fig:comparison_nb_games_videopinball}
    }

    \caption{Varying the number of games during the passive phase. The baseline curve corresponds to an expert initialized randomly.}
    \label{fig:comparison_nb_games}
\end{figure*}

\figurename \ref{fig:comparison_nb_games} compares the performance of the second active phase on different games when initialized by an AMN consolidated on subsets of 1, 2, 3 or 4 different Atari games.
In most games, the expert learns faster at the beginning of phase 2, compared to an agent trained without consolidation, but the number of games tackled during the first active phase does not seem to have any impact on the performance of the new experts.
The only exception is Breakout, where the effect of consolidation scales almost linearly depending on the number of games.
This goes to show that the expert can benefit from the learned features and  these features are more useful when they are extracted from multiple tasks.

\begin{figure}
  \centering
    \subfigure {
        \resizebox{0.36\textwidth}{!}{\includegraphics{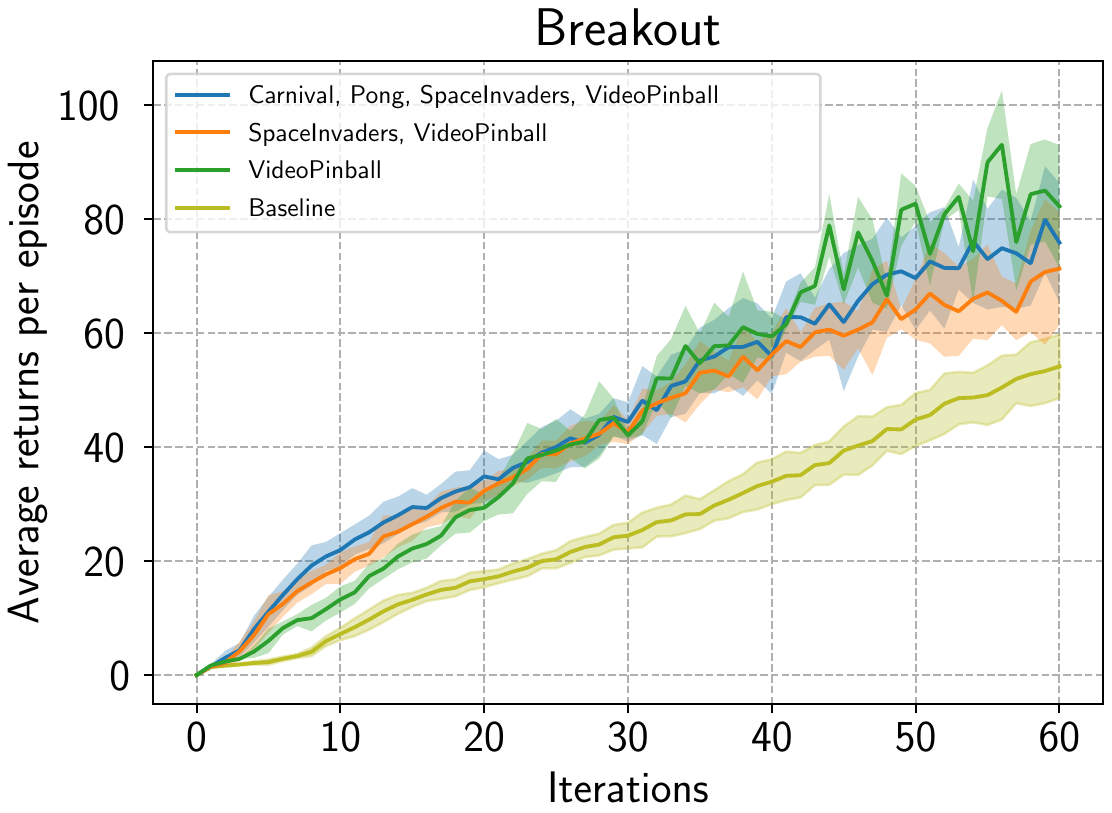}}
        \label{fig:comparison_with_pinball}
    }
    \subfigure {
        \resizebox{0.36\textwidth}{!}{\includegraphics{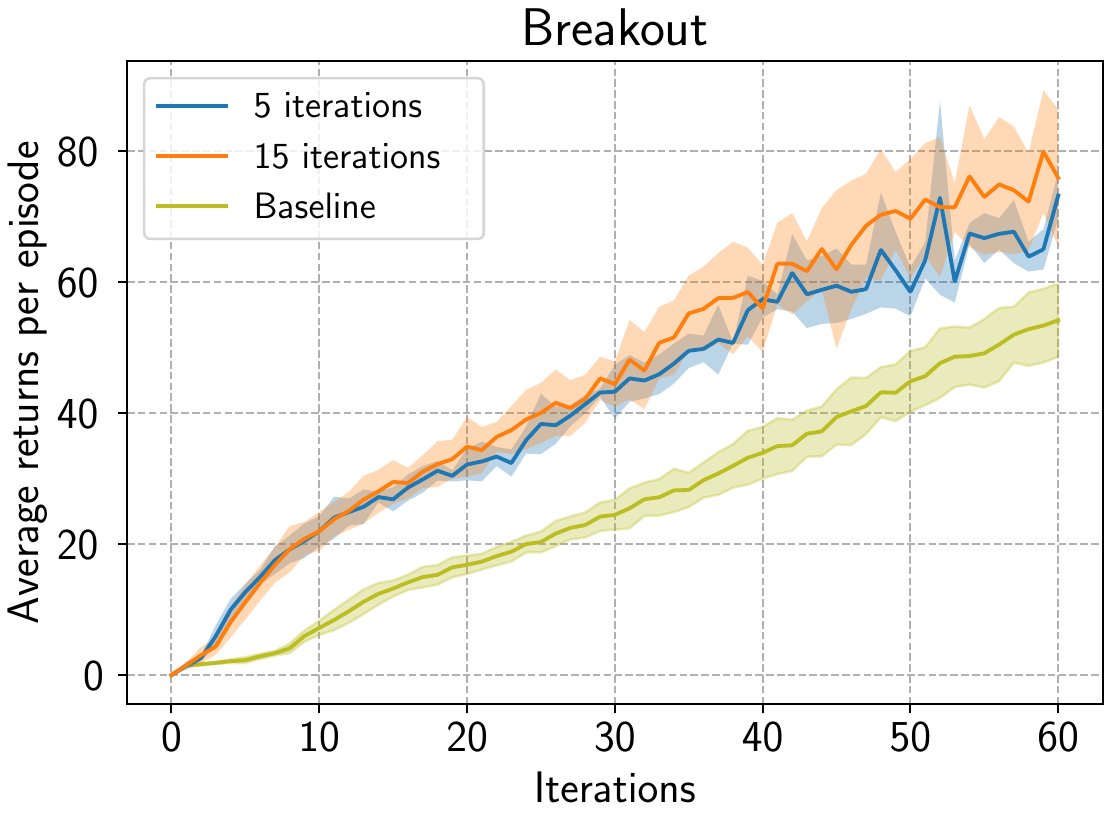}}
        \label{fig:comparison_shorter_night}
    }
    \caption{(Top) Transferring to Breakout from different sets of games. (Bottom) The AMN contains good features even without convergence in performance.}
    \label{fig:with_pinball_shorter_nights}
\end{figure}

In fact, the transfer effect appears to be more dependent on the games chosen for the consolidation step than on the number of games.
In the case of Breakout, only the presence of VideoPinball in the set of basis games has a significant effect on the newly trained experts.
\figurename \ref{fig:with_pinball_shorter_nights} (left) compares consolidation on three sets of games containing VideoPinball. In each case the Breakout expert reaches the same score with very little variation.

In the previous section, the final performance at the end of the passive phase was directly reusable by the new expert.
When transferring to new tasks, the policy is unlikely to provide any jumpstart in performance, therefore the passive phase final score is not as important as the features it managed to extract.
To measure the importance of the AMN score, we drastically shorten the passive phase to only 5 iterations instead of 15 (\figurename \ref{fig:with_pinball_shorter_nights}, right).
In that limited amount of time, the AMN is only able to reach 75\% of the VideoPinball expert performance, while it otherwise reaches around 105\%.
Despite this, the expert in the second active phase still reaches the same performance, confirming the improvement comes from a feature transfer as a good initialisation for optimisation, rather than from a direct policy transfer.

\subsection{Consolidation does not prevent transfer}

Complete results show that consolidation does not systematically induce improved learning speed or performance.
However, a consistent finding across experiments is that although increasing the number of games used during the passive phase does not necessarily lead to better performance, it also does not degrade it.
We conjecture that the consolidation process induces the same average improvement as the game which would have had the strongest impact if transferred alone.

\section{Transfer without consolidation}
\label{sec:transfer_without_consolidation}

\begin{figure*}[ht!]
    \centering
    \subfigure {
        \resizebox{0.31\textwidth}{!}{\includegraphics{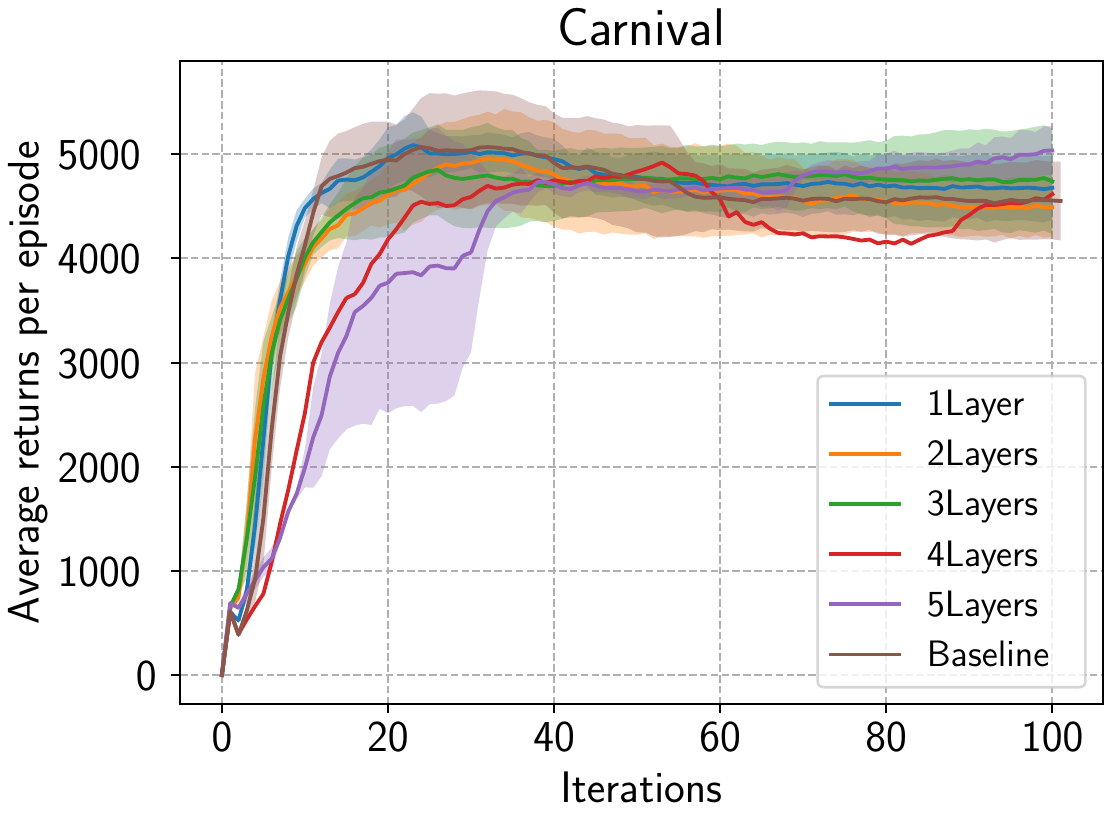}}
        \label{fig:transfer_carnival}
    }
    \subfigure {
        \resizebox{0.31\textwidth}{!}{\includegraphics{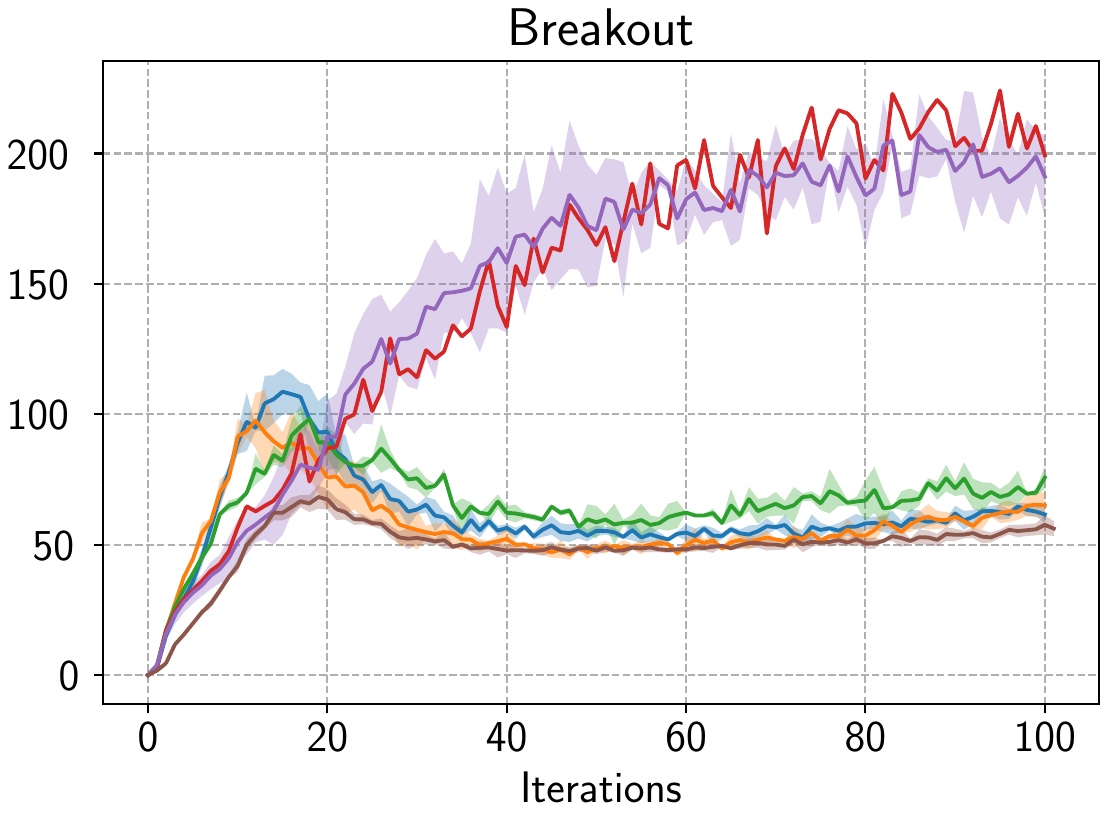}}
        \label{fig:transfer_breakout}
    }
    \subfigure {
        \resizebox{0.31\textwidth}{!}{\includegraphics{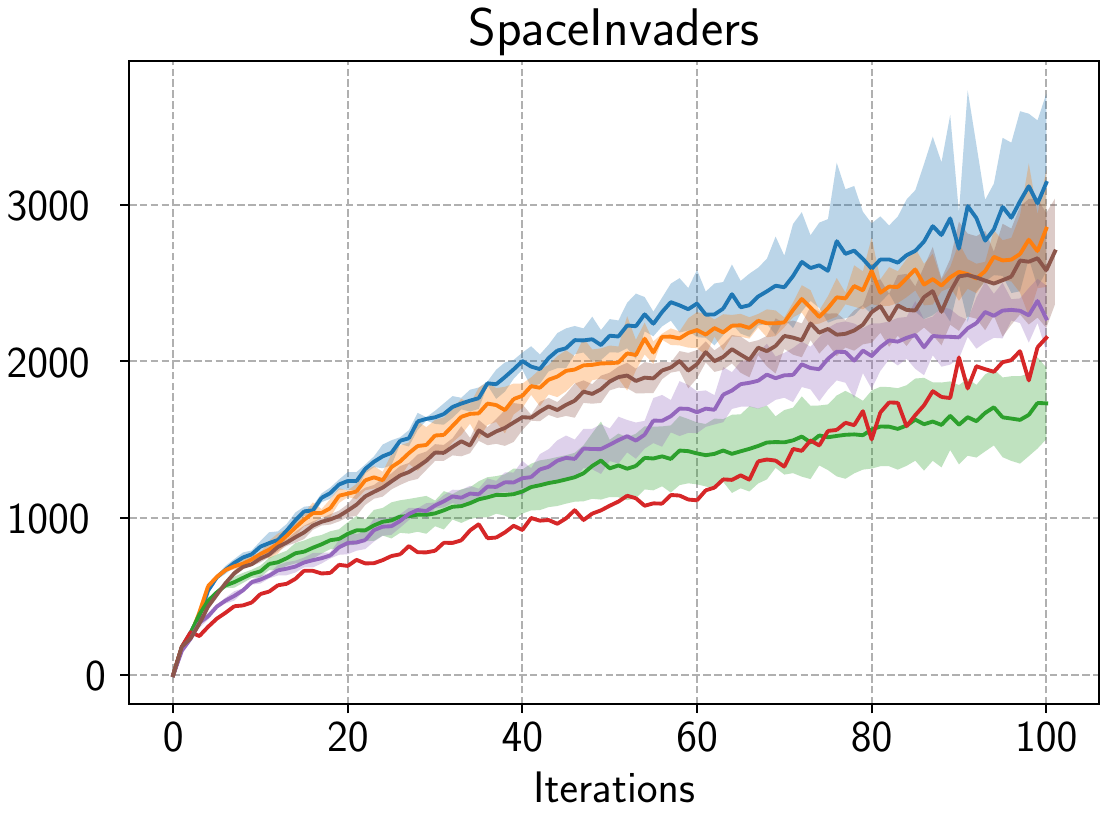}}
        \label{fig:transfer_spaceinvaders}
    }

    \caption{Transferring layers from an expert trained on VideoPinball. The baseline corresponds to an expert initialized randomly.}
    \label{fig:direct_transfer}
\end{figure*}

Since task-to-task transfer appears to have a significant importance, we factor out the distillation process and evaluate direct transfer of network weights between tasks without consolidation.
In this setting, an expert network is trained on a source task and then used as initialization for a new expert trained on a different target task.
This evaluation should provide a baseline confirming whether consolidation is useful at all when positive transfer is observed.
For a finer view on what precisely leads to positive transfer, we compare the performance obtained when only certain layers are transferred from an expert trained on VideoPinball, towards agents for other games (\figurename \ref{fig:direct_transfer}).
Recall that all networks here follow the 5-layered architecture introduced by \cite{mnih2015human}.
As expected, the effect of transfer is very task-dependent: either beneficial in terms of jumpstart, learning slope or asymptotic performance (positive transfer), or detrimental (negative transfer), compared to a baseline agent that is trained from scratch.
On Carnival, transferring one, two or three layers is equivalent to not transferring anything, whereas transferring the first four or five layers surprisingly deteriorates the very early training.
On the contrary, transferring any number of convolutional layers to Breakout does not have any effect on the performance, while transferring the features (first four layers) or the full policy (all layers) prevents the agent from plateauing and greatly improves the asymptotic performance.
Thus, the full transferred network seems to be a good starting point in parameters space to avoid local minima, but the visual features alone are not enough to fine-tune the network.
Finally, for SpaceInvaders, only the two first convolutional layers impact positively the agent's performance while the other layers have negative effects. This leads us to conjecture that only low-level visual features like edge detection are reusable in this case, and knowledge that is too specific only reduces the network plasticity.
Overall, these experiments confirm the popular assumption that negative transfer is as likely to occur than positive transfer with plain task-to-task network initialization. This sheds a different light on the conclusion of section \ref{sec:consolidation_for_transfer}. If a network trained on task A had a beneficial impact when transferred to task X, while a network trained on task B has a detrimental impact for task X, since the consolidation of the experts on A and B seem to retain the benefits inherited from A, consolidation appears as a means to filter out negative transfer while retaining the benefits of positive transfer.

\section{Conclusion}
\label{sec:conclu}

In this work, we propose an empirical discussion of the practical aspects of neural consolidation for knowledge transfer in neural networks, which we believe brings a new light on the matter for the community.
Investigating the necessary conditions for efficient consolidation in visual RL tasks pinpointed the strong dependence on the environment and hyperparameters.
We also highlighted that the appearance of useful features did not require convergence of the imitation process, and explored mechanisms that permit feature reuse after consolidation.
Our main finding is that consolidation appears to filter out negative transfers while maintaining positive transfers.
This result can be used \textit{as is} in a setting in which consolidation is already done.
In such settings it is good to notice that trying out transfer from all the source tasks to a new one can be done almost for free, in linear time, instead of trying all the transfers one by one.
More complex approaches can be imagined to discover transferability properties between sets of tasks (e.g. using a tree and dichotomy); these are left as future work.

\section*{Acknowledgements}
This work was granted access to the HPC resources of the CALMIP supercomputing center under allocation p21001.
The authors acknowledge the support of the French Defence Innovation Agency (AID) under grant ARAC.



\begin{thebibliography}{10}
\providecommand{\url}[1]{#1}
\csname url@samestyle\endcsname
\providecommand{\newblock}{\relax}
\providecommand{\bibinfo}[2]{#2}
\providecommand{\BIBentrySTDinterwordspacing}{\spaceskip=0pt\relax}
\providecommand{\BIBentryALTinterwordstretchfactor}{4}
\providecommand{\BIBentryALTinterwordspacing}{\spaceskip=\fontdimen2\font plus
\BIBentryALTinterwordstretchfactor\fontdimen3\font minus
  \fontdimen4\font\relax}
\providecommand{\BIBforeignlanguage}[2]{{%
\expandafter\ifx\csname l@#1\endcsname\relax
\typeout{** WARNING: IEEEtran.bst: No hyphenation pattern has been}%
\typeout{** loaded for the language `#1'. Using the pattern for}%
\typeout{** the default language instead.}%
\else
\language=\csname l@#1\endcsname
\fi
#2}}
\providecommand{\BIBdecl}{\relax}
\BIBdecl

\bibitem{sutton2018reinforcement}
R.~S. Sutton and A.~G. Barto, \emph{Reinforcement learning: An
  introduction}.\hskip 1em plus 0.5em minus 0.4em\relax MIT press, 2018.

\bibitem{pratt1991direct}
L.~Y. Pratt, J.~Mostow, and C.~A. Kamm, ``Direct transfer of learned
  information among neural networks,'' in \emph{AAAI}, 1991.

\bibitem{taylor2009transfer}
M.~E. Taylor and P.~Stone, ``Transfer {Learning} for {Reinforcement} {Learning}
  {Domains}: {A} {Survey},'' \emph{Journal of ML Research}, 2009.

\bibitem{rusu2016policy}
A.~A. Rusu, S.~G. Colmenarejo, {\c C}.~G{\"u}l{\c c}ehre, G.~Desjardins,
  J.~Kirkpatrick, R.~Pascanu, V.~Mnih, K.~Kavukcuoglu, and R.~Hadsell, ``Policy
  {Distillation},'' in \emph{ICLR}, 2016.

\bibitem{bucilua2006model}
C.~Bucilua, R.~Caruana, and A.~Niculescu-Mizil, ``Model compression,'' in
  \emph{SIGKDD Int. Conf. on Knowledge Discovery and Data Mining}, 2006.

\bibitem{hinton2015distilling}
G.~Hinton, O.~Vinyals, and J.~Dean, ``Distilling the {Knowledge} in a {Neural}
  {Network},'' in \emph{NeurIPS Deep Learning Workshop}, 2014.

\bibitem{parisotto2016actor-mimic}
E.~Parisotto, L.~J. Ba, and R.~Salakhutdinov, ``Actor-{Mimic}: {Deep}
  {Multitask} and {Transfer} {Reinforcement} {Learning},'' in \emph{ICLR},
  2016.

\bibitem{mcclelland1995why}
J.~L. McClelland, B.~L. McNaughton, and R.~C. O'Reilly, ``Why there are
  complementary learning systems in the hippocampus and neocortex: insights
  from the successes and failures of connectionist models of learning and
  memory.'' \emph{Psychological review}, 1995.

\bibitem{jung2016less-forgetting}
H.~Jung, J.~Ju, M.~Jung, and J.~Kim, ``Less-forgetting learning in deep neural
  networks,'' \emph{arXiv preprint arXiv:1607.00122}, 2016.

\bibitem{teh2017distral}
Y.~Teh, V.~Bapst, W.~M. Czarnecki, J.~Quan, J.~Kirkpatrick, R.~Hadsell,
  N.~Heess, and R.~Pascanu, ``Distral: Robust multitask reinforcement
  learning,'' in \emph{NeurIPS}, 2017.

\bibitem{parisi2019continual}
G.~I. Parisi, R.~Kemker, J.~L. Part, C.~Kanan, and S.~Wermter, ``Continual
  lifelong learning with neural networks: A review,'' \emph{Neural Networks},
  2019.

\bibitem{yoon2018lifelong}
J.~Yoon, E.~Yang, J.~Lee, and S.~J. Hwang, ``Lifelong {Learning} with
  {Dynamically} {Expandable} {Networks},'' in \emph{ICLR}, 2018.

\bibitem{rusu2016progressive}
A.~A. Rusu, N.~C. Rabinowitz, H.~Desjardins, G.and~Soyer, J.~Kirkpatrick,
  K.~Kavukcuoglu, R.~Pascanu, and R.~Hadsell, ``Progressive neural networks,''
  \emph{arXiv preprint arXiv:1606.04671}, 2016.

\bibitem{fernando2017pathnet}
C.~Fernando, D.~Banarse, C.~Blundell, Y.~Zwols, D.~Ha, A.~A. Rusu, A.~Pritzel,
  and D.~Wierstra, ``Pathnet: Evolution channels gradient descent in super
  neural networks,'' \emph{arXiv preprint arXiv:1701.08734}, 2017.

\bibitem{li2018learning}
Z.~Li and D.~Hoiem, ``Learning without {Forgetting},'' \emph{IEEE Transactions
  on Pattern Analysis and Machine Intelligence}, 2018.

\bibitem{kirkpatrick2017overcoming}
J.~Kirkpatrick, R.~Pascanu, N.~Rabinowitz, J.~Veness, G.~Desjardins, A.~A.
  Rusu, K.~Milan, J.~Quan, T.~Ramalho, A.~Grabska-Barwinska, D.~Hassabis,
  C.~Clopath, D.~Kumaran, and R.~Hadsell, ``Overcoming catastrophic forgetting
  in neural networks,'' \emph{Proc. of the National Academy of Sciences}, 2017.

\bibitem{zenke2017continual}
F.~Zenke, B.~Poole, and S.~Ganguli, ``Continual {Learning} {Through} {Synaptic}
  {Intelligence},'' in \emph{ICML}, 2017.

\bibitem{lopez-paz2017gradient}
D.~Lopez-Paz and M.~Ranzato, ``Gradient episodic memory for continual
  learning,'' in \emph{NeurIPS}, 2017.

\bibitem{rebuffi2017icarl}
S.-A. Rebuffi, A.~Kolesnikov, G.~Sperl, and C.~H. Lampert, ``{iCaRL}:
  {Incremental} {Classifier} and {Representation} {Learning},'' in \emph{IEEE
  CVPR}, 2017.

\bibitem{kaiser2020model}
{\L}.~Kaiser, M.~Babaeizadeh, P.~Miłos, B.~Osiński, R.~H. Campbell,
  K.~Czechowski, D.~Erhan, C.~Finn, P.~Kozakowski, S.~Levine, A.~Mohiuddin,
  R.~Sepassi, G.~Tucker, and H.~Michalewski, ``Model {Based} {Reinforcement}
  {Learning} for {Atari},'' in \emph{ICLR}, 2020.

\bibitem{ha2018world}
D.~Ha and J.~Schmidhuber, ``World models,'' \emph{arXiv preprint
  arXiv:1803.10122}, 2018.

\bibitem{shin2017continual}
H.~Shin, J.~K. Lee, J.~Kim, and J.~Kim, ``Continual learning with deep
  generative replay,'' in \emph{NeurIPS}, 2017.

\bibitem{berseth2018progressive}
G.~Berseth, C.~Xie, P.~Cernek, and M.~V. de~Panne, ``Progressive reinforcement
  learning with distillation for multi-skilled motion control,'' in
  \emph{ICLR}, 2018.

\bibitem{schwarz2018progress}
J.~Schwarz, W.~Czarnecki, J.~Luketina, A.~Grabska-Barwinska, Y.~W. Teh,
  R.~Pascanu, and R.~Hadsell, ``Progress \& compress: A scalable framework for
  continual learning,'' in \emph{ICML}, 2018.

\bibitem{blakeman2020complementary}
S.~Blakeman and D.~Mareschal, ``A complementary learning systems approach to
  temporal difference learning,'' \emph{Neural Networks}, 2020.

\bibitem{finn2017model-agnostic}
C.~Finn, P.~Abbeel, and S.~Levine, ``Model-{Agnostic} {Meta}-{Learning} for
  {Fast} {Adaptation} of {Deep} {Networks},'' in \emph{ICML}, 2017.

\bibitem{beaulieu2020learning}
S.~Beaulieu, L.~Frati, T.~Miconi, J.~Lehman, K.~O. Stanley, J.~Clune, and
  N.~Cheney, ``Learning to {Continually} {Learn},'' in \emph{ECAI}, 2020.

\bibitem{he2020task}
X.~He, J.~Sygnowski, A.~Galashov, A.~A. Rusu, Y.~W. Teh, and R.~Pascanu, ``Task
  {Agnostic} {Continual} {Learning} via {Meta} {Learning},'' in \emph{{ICML}
  {Workshop} on {LifelongML}}, 2020.

\bibitem{hessel2018rainbow}
M.~Hessel, J.~Modayil, H.~van Hasselt, T.~Schaul, G.~Ostrovski, W.~Dabney,
  D.~Horgan, B.~Piot, M.~G. Azar, and D.~Silver, ``Rainbow: Combining
  improvements in deep reinforcement learning,'' in \emph{AAAI}, 2018.

\bibitem{castro2018dopamine}
P.~S. Castro, S.~Moitra, C.~Gelada, S.~Kumar, and M.~G. Bellemare, ``Dopamine:
  A research framework for deep reinforcement learning,'' \emph{arXiv preprint
  arXiv:1812.06110}, 2018.

\bibitem{mnih2015human}
V.~Mnih, K.~Kavukcuoglu, D.~Silver, A.~A. Rusu, J.~Veness, M.~G. Bellemare,
  A.~Graves, M.~Riedmiller, A.~K. Fidjeland, G.~Ostrovski \emph{et~al.},
  ``Human-level control through deep reinforcement learning,'' \emph{nature},
  2015.

\bibitem{schaul2015prioritized}
T.~Schaul, J.~Quan, I.~Antonoglou, and D.~Silver, ``Prioritized experience
  replay,'' \emph{arXiv preprint arXiv:1511.05952}, 2015.

\bibitem{bellemare2013arcade}
M.~G. Bellemare, Y.~Naddaf, J.~Veness, and M.~Bowling, ``The {Arcade}
  {Learning} {Environment}: {An} {Evaluation} {Platform} for {General}
  {Agents},'' \emph{Journal of Artificial Intelligence Research}, 2013.

\bibitem{yu2020gradient}
T.~Yu, S.~Kumar, A.~Gupta, S.~Levine, K.~Hausman, and C.~Finn, ``Gradient
  surgery for multi-task learning,'' in \emph{NeurIPS}, 2020.

\end{thebibliography}
\end{document}